\documentclass[submission,copyright,creativecommons]{eptcs}
\providecommand{\event}{ARCADE 2019} % Name of the event you are submitting to
\usepackage{breakurl}             % Not needed if you use pdflatex only.

\title{On Quantified Modal Theorem Proving for Modeling Ethics}%

\author{Naveen Sundar Govindarajulu
\institute{Rensselaer AI \& Reasoning Lab\\ Troy, New York}
\email{naveensundarg@gmail.com}
\and
Selmer Bringsjord
\institute{Rensselaer AI \& Reasoning Lab \\Department of Cognitive
  Science \\ Department of Computer Science \\Rensselaer Polytechnic Institute\\
Troy, New York}
\email{\quad selmer.bringsjord@gmail.com }
\and
Matthew Peveler
\institute{Rensselaer AI \& Reasoning Lab \\ Department of Computer Science \\Rensselaer Polytechnic Institute\\
Troy, New York}
\email{matt.peveler@gmail.com}
}
\def\titlerunning{Theorem Proving for Quantified Modal Logics}
\def\authorrunning{Govindarajulu, Bringsjord, and Peveler}

\usepackage{mdframed}
\usepackage{graphicx}
\usepackage{verbatim}
\usepackage{fancybox}
\usepackage{amssymb}
\usepackage{amsmath}
\usepackage{proof}  
\usepackage{bussproofs}
\usepackage{verbatim}
\usepackage{algorithm2e}
\usepackage{hyperref}
\hypersetup{
  colorlinks,
  citecolor=blue,
  filecolor=blue,
  linkcolor=blue,
  urlcolor=blue
}
\usepackage{hyperref}
\usepackage{paralist}
\usepackage[]{algorithm2e}
\usepackage{graphicx}
\usepackage{subcaption}
\usepackage{booktabs,arydshln}
\usepackage{listings}

\usepackage{xcolor}
 \usepackage{booktabs}

\definecolor{codegreen}{rgb}{0,0.6,0}
\definecolor{codegray}{rgb}{0.5,0.5,0.5}
\definecolor{codepurple}{rgb}{0.58,0,0.82}
\definecolor{backcolour}{rgb}{0.95,0.95,0.95}
 
\lstdefinestyle{codestyle}{
    backgroundcolor=\color{backcolour},   
    commentstyle=\color{codegreen},
    keywordstyle=\color{blue},
    numberstyle=\tiny\color{codegray},
    stringstyle=\color{codepurple},
    basicstyle=\ttfamily\footnotesize,
    breakatwhitespace=false,         
    breaklines=true,                 
    captionpos=b,                    
    keepspaces=true,                 
    numbers=left,                    
    numbersep=5pt,                  
    showspaces=false,                
    showstringspaces=false,
    showtabs=false,                  
    tabsize=1
}
\makeatletter
\def\adl@drawiv#1#2#3{%
        \hskip.35\tabcolsep
        \xleaders#3{#2.5\@tempdimb #1{1}#2.5\@tempdimb}%
                #2\z@ plus1fil minus1fil\relax
        \hskip.5\tabcolsep}
\newcommand{\cdashlinelr}[1]{%
  \noalign{\vskip\aboverulesep
           \global\let\@dashdrawstore\adl@draw
           \global\let\adl@draw\adl@drawiv}
  \cdashline{#1}[.4pt/1pt]
  \noalign{\global\let\adl@draw\@dashdrawstore
           \vskip\belowrulesep}}
\makeatother

\usepackage{tabu}

\newcommand{\cc}{\ensuremath{\mathscr{C}}}

\usepackage{mathrsfs}
\usepackage{pslatex}
\usepackage{booktabs}
\usepackage{verbatim}
\usepackage{color}
\usepackage{amsfonts}
\usepackage{amsmath}
\usepackage{amssymb}

\usepackage{xcolor}

\usepackage{proof}
\usepackage{fancybox}
\usepackage{bussproofs}

\mathchardef\mhyphen="2D

\newcommand{\lsort}[1]{%
  \ensuremath{\mbox{\textsf{#1}}}}
\newcommand{\defsort}[2]{%
  \newcommand{#1}{\lsort{#2}}}

\defsort{\Action}{Action}
\defsort{\Time}{Time}
\defsort{\Self}{Self}
\defsort{\SortName}{SortName}

\defsort{\Agent}{Agent}
\defsort{\Entrant}{Entrant}
\defsort{\ActionType}{ActionType}
\defsort{\Moment}{Moment}
\defsort{\Boolean}{Boolean}
\defsort{\PayOut}{PayOut}
\defsort{\Fluent}{Fluent}
\defsort{\Event}{Event}
\defsort{\Object}{Object}
\defsort{\Numeric}{Numeric}

\newcommand{\lsymbol}[1]{%
  \ensuremath{\mathit{#1}}}
\newcommand{\defsymbol}[2]{%
  \newcommand{#1}{\lsymbol{#2}}}
\defsymbol{\action}{action}
\defsymbol{\initially}{initially}
\defsymbol{\holds}{holds}
\defsymbol{\mirrored}{mirrored}
\defsymbol{\happens}{happens}
\defsymbol{\clipped}{clipped}
\defsymbol{\initiates}{initiates}
\defsymbol{\terminates}{terminates}
\defsymbol{\prior}{prior}
\defsymbol{\interval}{interval}
\defsymbol{\refrain}{refrain}
\defsymbol{\harm}{harm}

\defsymbol{\does}{does}
\defsymbol{\plans}{plans}
\defsymbol{\act}{act}
\defsymbol{\react}{react}
\defsymbol{\payTot}{pay_{tot}}
\defsymbol{\fight}{fight}
\defsymbol{\coop}{coop}
\defsymbol{\enter}{enter}
\defsymbol{\stayout}{stayout}
\defsymbol{\learns}{learns}
\defsymbol{\redsplotched}{red\mhyphen splotched}
\defsymbol{\hassplotch}{has\mhyphen splotch}
\defsymbol{\payoff}{payoff}

\defsymbol{\capable}{capable}
\defsymbol{\destroyed}{destroyed}

\defsymbol{\removesplotch}{remove\mhyphen splotch}
\defsymbol{\remove}{remove}

\defsymbol{\cogito}{cogito}
\defsymbol{\named}{named}
\defsymbol{\deter}{deter}
\defsymbol{\enhance}{enhance}
\defsymbol{\attack}{attack}
\defsymbol{\cost}{cost}

\newcommand{\lconstant}[1]{%
  \ensuremath{\mbox{\textsf{#1}}}}
\newcommand{\defconstant}[2]{%
  \newcommand{#1}{\lconstant{#2}}}
\defconstant{\Enter}{Enter}
\defconstant{\StayOut}{StayOut}
\defconstant{\Fight}{Fight}
\defconstant{\Acquiesce}{Acquiesce}
\defconstant{\cs}{cs }
\defconstant{\estimate}{estimate}

\newcommand{\lmodality}[1]{%
  \ensuremath{\mathbf{#1}}}
\newcommand{\defmodality}[2]{%
  \newcommand{#1}{\lmodality{#2}}}
\defmodality{\common}{C}
\defmodality{\knows}{K}
\defmodality{\believes}{B}
\defmodality{\perceives}{P}
\defmodality{\mental}{M}
\defmodality{\ought}{O}
\defmodality{\goal}{G}

\defmodality{\desires}{D}
\defmodality{\intends}{I}
\defmodality{\says}{S}

\newcommand{\DCEC}{\ensuremath{\mathcal{DCEC}}}

\newcommand{\DDE}{\ensuremath{\mathcal{DDE}}}
\newcommand{\DDES}{\ensuremath{\mathcal{DDE}^\ast}}

\newcommand{\sep}{\ \lvert \ }

\defsymbol{\wipe}{wipe\mhyphen fore\mhyphen head}

\defsymbol{\us}{\mathsf{us}}
\defsymbol{\iran}{\mathsf{iran}}
\defsymbol{\israel}{\mathsf{israel}}
\defsymbol{\russia}{\mathsf{russia}}
\defsymbol{\T}{\mathsf{T}}
\defsymbol{\ugv}{\mathsf{ugv}}
\defsymbol{\uav}{\mathsf{uav}}
\defsymbol{\carrying}{carrying}
\defsymbol{\firefight}{firefight}
 \defsymbol{\mission}{mission}
\defconstant{\silence}{silence}
\defconstant{\now}{now}

\defconstant{\solda}{\ensuremath{\mathsf{soldier_A}}}
\defconstant{\soldb}{\ensuremath{\mathsf{soldier_B}}}
\defconstant{\sold}{\ensuremath{\mathsf{soldier}}}
\defconstant{\commander}{\ensuremath{\mathsf{commander}}}

\defconstant{\enemyterritory}{enemyterritory}
\defconstant{\baseb}{\ensuremath{\mathsf{base_b}}}
\defsymbol{\move}{move}
\defsymbol{\allowed}{\mathbf{allowed}}
\defconstant{\past}{past}
\defconstant{\jack}{jack}

\defsymbol{\rich}{rich}

\let\originalleft\left
\let\originalright\right
\renewcommand{\left}{\!\mathopen{}\mathclose\bgroup\originalleft}
\renewcommand{\right}{\aftergroup\egroup\!\originalright}

\defconstant{\robot}{Robot}

\newcounter{parens}
\def\countlparen{%
    \addtocounter{parens}{1} \overset{\ensuremath{{\color[rgb]{0.7,.7,0.7} ^{\the\value{parens}}}
}}{\lparen}%
}
\def\countrparen{%
\overset{\ensuremath{{\color[rgb]{0.7,.7,0.7} ^{\the\value{parens}}}
}}{\rparen}\addtocounter{parens}{-1}%
}
\let\lparen(
\let\rparen)
\begingroup
    \catcode`(\active
    \catcode`)\active
    \gdef\countparens{%
        \let(\countlparen
        \let)\countrparen
    }
\endgroup   
\newenvironment{nested parentheses}
{%
    \catcode`(\active
    \catcode`)\active
    \countparens
    \setcounter{parens}{0}%
}

\newcommand{\unify}{\mathsf{unify}}

\begin{document}
\maketitle

\begin{abstract}
  In the last decade, formal logics have been used to model a wide
  range of ethical theories and principles with the goal of using
  these models within autonomous systems.  Logics for modeling ethical
  theories, and their automated reasoners, have requirements that are
  different from modal logics used for other purposes, e.g.\ for
  temporal reasoning.  Meeting these requirements necessitates
  investigation of new approaches for proof automation.  Particularly,
  a quantified modal logic, the \textbf{deontic cognitive event
    calculus} (\DCEC), has been used to model various versions of the
  doctrine of double effect, \textit{akrasia}, and virtue ethics.
  Using a fragment of \DCEC, we outline these distinct characteristics
  and present a sketches of an algorithm that can help with some
  aspects proof automation for \DCEC.
\end{abstract}

\section{Introduction}
Modal logics have been used for decades to model and study a diverse
set of subjects --- e.g.\ temporal reasoning, multi-agent systems,
linguistic content and phenomena, and game theory \cite[Part
  4]{blackburn2006handbook}.  While deontic modal logics have been
used to study ethical principles, it is only recently that such logics
have been considered in a rigorous manner \cite{sb_etal_ieee_robots}
with the goal of either using them in a computational system or using
such a logic to analyze computational systems.

For example, a quantified modal logic, the \textbf{deontic cognitive
  event calculus} (\DCEC), has been used recently to model various
versions of the \emph{Doctrine of Double Effect}, \emph{akrasia}, and
\emph{virtue ethics}
\cite{govindarajuluetalaies2019,DBLP:journals/corr/abs-1805-07797,nsg_sb_dde_ijcai,Govindarajulu2019SelfSacrifice,akratic_robots_ieee_n}.
These ethical principles and theories have a unique set of
characteristics when compared with other domains, e.g.\ with temporal
reasoning, in which modal logics have been used.\footnote{A note on
  the terms ``ethical principles'' and ``ethical theories.''  An
  ethical theory is generally broader and more fundamental than an
  ethical principle.  An ethical principle is ultimately a declarative
  statement usually cast under one or more ethical theories.  E.g.,
  the principle that one ought always to act with the intention to
  maximize utility for everyone would fall under the ethical theory
  known as \textit{utilitarianism}.  For a classic presentation of the
  main ethical theories and their key principles, see
  \cite{feldman_introductory_ethics}.} This implies that logics for
modeling ethical theories have requirements that are different than
those for modal logics used for other purposes.  These requirements
dictate investigation of new approaches for proof automation.  We
present a central set of these requirements in this paper.  Using a
fragment $\cc^1 $ of \DCEC, we also present an algorithm that can help
enable proof automation which partially satisfies these requirements.
% During the workshop, we will compare and contrast with existing
% systems by Benzm\"{u}ller and colleagues
% \cite{ecai2014_paleo_godel_proof,inconsistency_godel_modal_ijcai2016}
% and present initial results on benchmarking the above two algorithms
% and their implementations.

%%% Local Variables:
%%% mode: latex
%%% TeX-master: "main"
%%% End:

%% SB@@@

\section{Requirements for Modeling Ethical Theories}

To illustrate the unique characteristics required for modeling ethical
theories and principle, we use the \textbf{Doctrine of Double Effect}
%% NAVEEN: I think we need to consistentyl capitalize mention of DDE
%% when spelled out.  Have tried to do that.  Also, in intro.tex I
%% believe the phrase is italicized, not bolded.  First occurrence
%% should be bolded, I think.  //S
%% SELMER: Good catch. Will do so
(\DDE) augmented to handle self sacrifice.  \DDE\ is an ethical
principle that can account for human judgment in \textbf{moral
  dilemmas}: situations in which all available actions have both
significantly good \textit{and} significantly bad consequences.
According to \DDE, an action $\alpha$ in such a situation is
permissible \emph{iff} 
%% NAVEEN: Do you really want the --- in the previous line?  //S
%% SELMER: removed

\begin{quote}``\begin{inparaenum}[(1)] \item it is
  morally neutral; \item the net good consequences outweigh the bad
  consequences by a large amount; and \item some of the good
  consequences are intended, while none of the bad consequences
  are. \end{inparaenum} \cite{nsg_sb_dde_2017}'' \end{quote}

\noindent A formalization of \DDE\ is presented in
\cite{nsg_sb_dde_2017}.  While \DDE\ has some empirical support
\cite{cushman2006role}, it cannot account for instances of
self-sacrifice.  To handle self-sacrifice, an augmented version,
$\DDES$, is presented and formalized in
\cite{Govindarajulu2019SelfSacrifice}.  We now present an informal
version of $\DDES$ to illustrate the requirements.  We assume
%% NAVEEN:  Here, \DDE isn't used.  Is this correct?  //S
%% SELMER: Yes, this is correct. Refering to the self-sacrifice version of DDE.
there is an ethical hierarchy of actions (e.g.\ \textit{forbidden},
\textit{morally neutral}, \textit{obligatory}); see
\cite{bringsjord201721st}.  We also assume that we have a utility or
goodness function for states of the world or effects.  For an
autonomous agent $a$, an action $\alpha$ in a situation $\sigma$ at
time $t$ is said to be $\DDES$-compliant \emph{iff}:
%% NAVEEN:  Here, \DDE isn't used.  Is this correct?  //S
%% SELMER: Yes, this is correct. Refering to the self-sacrifice version of DDE.

%% NAVEEN: In the following enumeration, I'm not sure what your
%% underlying rationale is for bolding text or not bolding text, so I
%% didn't touch anything along this line.  //S
%% SELMER: Removed the bolded text.
\begin{small}
\begin{enumerate}
\item[$\mathbf{C}_1$] At the time of the action, the agent $a$
  executing the action {believes} that the action is not
  forbidden (where, again, we assume an ethical hierarchy such as the
  one given by Bringsjord \cite{bringsjord201721st}, and require that
  the action be morally neutral or above morally neutral in such a
  hierarchy);
\item[$\mathbf{C}_2$] At the time of the action, the agent $a$
  {believes} that the net utility or goodness of the action is
  greater than some positive amount $\gamma$;
\item[$\mathbf{C}_{3a}$] At the time of the action, the agent $a$
  performing the action {intends} only the good effects;
\item[$\mathbf{C}_{3b}$] At the time of the action, the agent $a$ does
  not {intend} any of the bad effects;
\item[$\mathbf{C}_4$] the bad effects are not used by $a$ as a means
  to obtain the good effects {[unless $a$ knows that the bad
      effects are confined to only $a$ itself]}; and
\item[$\mathbf{C}_5$] if there are bad effects, the agent would rather
  the situation be different and the agent not have to perform the
  action; that is, the action is unavoidable.
\end{enumerate}
\end{small}
%\vspace{-0.2in}

\noindent With $\DDES$ as the background, we outline the following
%% NAVEEN:  \DDE?
%% SELMER: Yes, this is correct. Refering to the self-sacrifice version of DDE.
requirements that are necessary in modeling not only $\DDES$ but
also other ethical theories and principles, such as virtue ethics and
\textit{akrasia}.  We split the requirements into two parts:
requirements for the logic, and additional requirements for the
reasoner.
 
\begin{mdframed}[linecolor=white, nobreak, frametitle= Requirements for the Logic, frametitlebackgroundcolor=gray!07, backgroundcolor=gray!03, roundcorner=8pt]
\begin{enumerate}

\item[$\mathbf{R}_1$] \textbf{Multiple Modalities}: Ethical principles
  have statements that take into account an agent's beliefs,
  intentions, obligations, etc.  Any acceptable logic should be able
  to handle this.

 \item[$\mathbf{R}_2$] \textbf{Time-Indexed Modalities}: Intentions
   and beliefs at the time of an action matter rather than intentions
   and beliefs at other times.

\item[$\mathbf{R}_3$] \textbf{\emph{De se} Agent Modalities}
  Agent-indexed modalities are common in BDI (belief/desire/intention)
  logics \cite{wooldridge.mas}, but $\mathbf{C}_4$ requires self
  beliefs known as \emph{de se} beliefs.  This requires modalities
  indexed by \emph{de se} agents.  This is needed to model statements
  such as \emph{``a believes that a herself believes that ...''}.  For
  more details on \emph{de se} beliefs, please see
  \cite{Govindarajulu2019SelfSacrifice} and \cite{mgmmm_ptai_sb}.

\item[$\mathbf{R}_4$] \textbf{Quantifiers}: Quantifiers are needed to
  handle comparisons between actions and for ordering actions by their
  consequences.

\end{enumerate}
\end{mdframed}

\noindent While the above core requriments are needed for $\DDES$,
%% NAVEEN: \DDE?  Define a new new command as needed?  //S
other features, such as the ability to represent uncertainty and
counterfactuals, may be needed for some ethical theories.  We omit
these requirements from the core list above as there has not been as
much discussion around these features.  In addition to handling the
above requirements, any reasoner for the logic should have the
following capabilities:

\begin{mdframed}[linecolor=white, frametitle= Additional Requirements
  for the Reasoner, frametitlebackgroundcolor=gray!07, backgroundcolor=gray!03, nobreak, roundcorner=8pt]

\begin{enumerate}
\item \textbf{Builtin Theories}: Handling of simple arithmetic and
  causation.  This is required for efficiently computing consequences,
  and causes of actions.
\item \textbf{Justifications and Explanations}: Any reasoning system
  in an ethically charged scenario should be able to explain and
  present its reasoning in a verifiable and understandable manner.

\item \textbf{Answer Finding}: The reasoner should not only be used
  for proving that an action is ethical but should also be capable of
  finding the most ethical action in a given situation.
\end{enumerate}
\end{mdframed}

%%% Local Variables:
%%% mode: latex
%%% TeX-master: "main"
%%% End:

\section{A Sparse Calculus $\cc^1$}

$\cc^1$ is a straightforward modal extension of first-order logic that
satisfies $\mathbf{R}_1$, $\mathbf{R}_2$, and $\mathbf{R}_4$.  We have
a modal operator $\believes$ for belief, an operator $\ought$ for
obligation, and $\mathbf{G}$ to denote goals.  The syntax and
inference schemata of the system are shown below.  Assume that we have
a first-order alphabet augmented with a fixed finite set of symbols
for agents $Ag=\{a_1, \ldots, a_n\}$ and a set of totally ordered
symbols for time $T = \langle t_0, \ldots, t_n, \ldots \rangle$.
Sometimes we use $a$ for $a_i$ and $t$ for $t_i$.  $\phi$ is a
meta-variable for formulae, and $A$ is any first-order atomic symbol.
Given this, the grammar for wffs of $\cc^1$ follows.
%$\{x, y, z,\ldots \}$ are first-order variables.

 \begin{small}

\begin{equation*}
 \begin{aligned}
    \mathit{s_i}&::=  \mbox{standard first-order terms}\\
    \mathit{\phi}&::= \left\{ 
    \begin{aligned}
     &A(s_1, \ldots s_n) \\
&  \neg \phi \sep \phi \lor 
     \psi \sep \forall x. \phi% \sep \phi \land \psi \sep \phi \land \psi  \sep \phi \rightarrow \psi \sep \phi \leftrightarrow \psi
\\     %\sep \forall x. \phi\\ 
     &   \believes (a,t, \phi)  \sep \ought(a, t, \phi, \psi) \sep \goal(a, t,\psi)  %\sep \desires(a,t,\phi) \sep \intends(a,t,\phi) \sep \ought(a,t,\phi, \psi) 
      \end{aligned}\right.
  \end{aligned}
\end{equation*}
\end{small}
  % $\believes(a,\phi)$ stands for agent $a$ believing $\phi$. The syntax
% lets us formalize statements of the form \emph{``Jack believes that
%   Jill believes
%   that it is snowing.''}  One formalization could be:
%  \begin{equation*}
%  \begin{aligned}
% \believes\Big(\mathit{jack},  
% \believes\big(\mathit{jill},  \holds(snowing, t)\big)\Big)  
%   \end{aligned}
% \end{equation*}

\noindent $\believes (a,t, \phi)$ states that $a$ believes at time $t$
that $\phi$ holds.  $\ought (a,t, \phi, \psi)$ states that $a$ ought
to $\psi$ at time $t$ that if $\phi$ holds.  $\goal(a, t,\psi)$ states
that $a$ has as a goal $\psi$ at time $t$.

\paragraph{Inference System} 

We have three inference schemata: $\{I_{R}, I_B, I_O\}$, shown in
Table 1.  $\unify(a, b)$ denotes the most general first-order unifier
of $a$ and $b$.  First-order reasoning is performed through $I_R$,
which is just first-order resolution.  Reasoning with beliefs is done
with $I_B$.  Beliefs propagate forward in time.  Reasoning with
obligations is handled with $I_O$: If an agent believes it has an
obligation $\psi$ when $\phi$, and believes that $\phi$, then it has a
goal $\psi$.

\begin{footnotesize}
\begin{center}
\begin{tabular}{p{1.5cm}c}  
\toprule
\textbf{Description}    & \textbf{Inference Scheme} \\
\midrule
$I_{R}$ & \begin{footnotesize}$\infer[{[I_{\mathbf{Res}}]}]{(\phi_1 \lor \ldots \lor \ldots \phi_n
    \lor \psi_1\lor \ldots \lor \psi_m)\theta}{\phi_1 \lor \ldots \chi
    \ldots\lor\phi_n  \hspace{40pt}\ \psi_1 \lor \ldots \lnot
    \mathbf{\chi}'  \ldots \lor \psi_m \hspace{10pt} \mbox{where } \theta = \unify(\chi, \chi')}$\end{footnotesize} \\
\addlinespace[1em]\cdashlinelr{1-2}\addlinespace[1em]
$I_B$   &  \begin{footnotesize}$\infer[{[I_{\mathbf{B}}]}]{\believes\big(a, t (\phi_1 \lor \ldots \lor \ldots \phi_n
    \lor \psi_1\lor \ldots \lor \psi_m) \theta\big)}{\believes(a, t_1,
    \phi_1 \lor \ldots \chi
    \ldots\lor\phi_n) \hspace{15pt}\ \believes(a,  t_2, \psi_1 \lor \ldots \lnot
    \mathbf{\chi}'  \ldots \lor \psi_m) \hspace{10pt} \mbox{where }  t
    \geq t_1, t_2; \mbox{ and }\theta = \unify(\chi, \chi')}$ \end{footnotesize}\\
\addlinespace[1em]\cdashlinelr{1-2}\addlinespace[1em]
$I_O$ &  \begin{footnotesize}$\infer[{[I_{\mathbf{O}}]}]{ \goal(a, \psi)}{\believes(a, \phi) \hspace{20pt}  \believes(a, \ought(a, \phi, \psi)) }$ \end{footnotesize}\\
 \bottomrule
\end{tabular}
\end{center}
\end{footnotesize}

\paragraph{Proof from $\Gamma$ to
    $\phi$:}
  A proof $\Pi^\Gamma_\phi$ from $\Gamma$ to $\phi$ consists of a
  sequence of formulae $\phi_1, \phi_2, \ldots, \phi_n$ such that
  \begin{inparaenum}[(i)] \item $\phi_n \equiv \phi$; and \item for
    all $1\leq i<n$, $\phi_i$ is derived from $\{\phi_j \lvert j <
    i\}$ using $I_\mathbf{R}$, $I_{\mathbf{B}}$, or
    $I_{\mathbf{O}}$. \end{inparaenum} $\Gamma\vdash\phi$ denotes that
  there is a proof $\Pi^\Gamma_\phi$ from $\Gamma$ to $\phi$. 
% Note that while the syntax and inference schemata are specified
% using just $\lor$ $$, we will use other logical connectives $\land,
% \rightarrow$ to make the presentation below easier to read.

%%% Local Variables:
%%% mode: latex
%%% TeX-master: "main"
%%% End:

\section{An Algorithm Sketch}
We now present an algorithm for handling the proof system for $\cc^1$.
Our goal is to leverage advances in first-order theorem proving to
build the relevant reasoner.  There are two straightforward but flawed
ways this can be done.  In the first approach, modal operators are
simply represented by first-order predicates.  This approach is the
fastest but can quickly lead to well-known inconsistencies, as
demonstrated in \cite{selmer_naveen_metaphil_web_intelligence}.  In
the second approach, the entire proof theory is implemented
intricately in first-order logic, and the reasoning is carried out
within first-order logic.  Here, the first-order theorem prover simply
functions as a declarative programming system.  This approach, while
accurate, can be inefficient.

Our algorithm is based on a technique we term \textbf{shadowing}.  At
at high-level, we alternate between calling a first-order theorem
prover and applying modal inference schemata.  When we call the
first-order prover, all modal atoms are converted into propositional
atoms (i.e.\ the former are shadowed), to prevent substitution into
modal contexts.  This approach achieves speed without sacrificing
consistency.  The algorithm is briefly described below.

\begin{figure}
    \centering
    \begin{subfigure}[b]{0.5\textwidth}
      \shadowbox{
        \includegraphics[width=.90\linewidth]{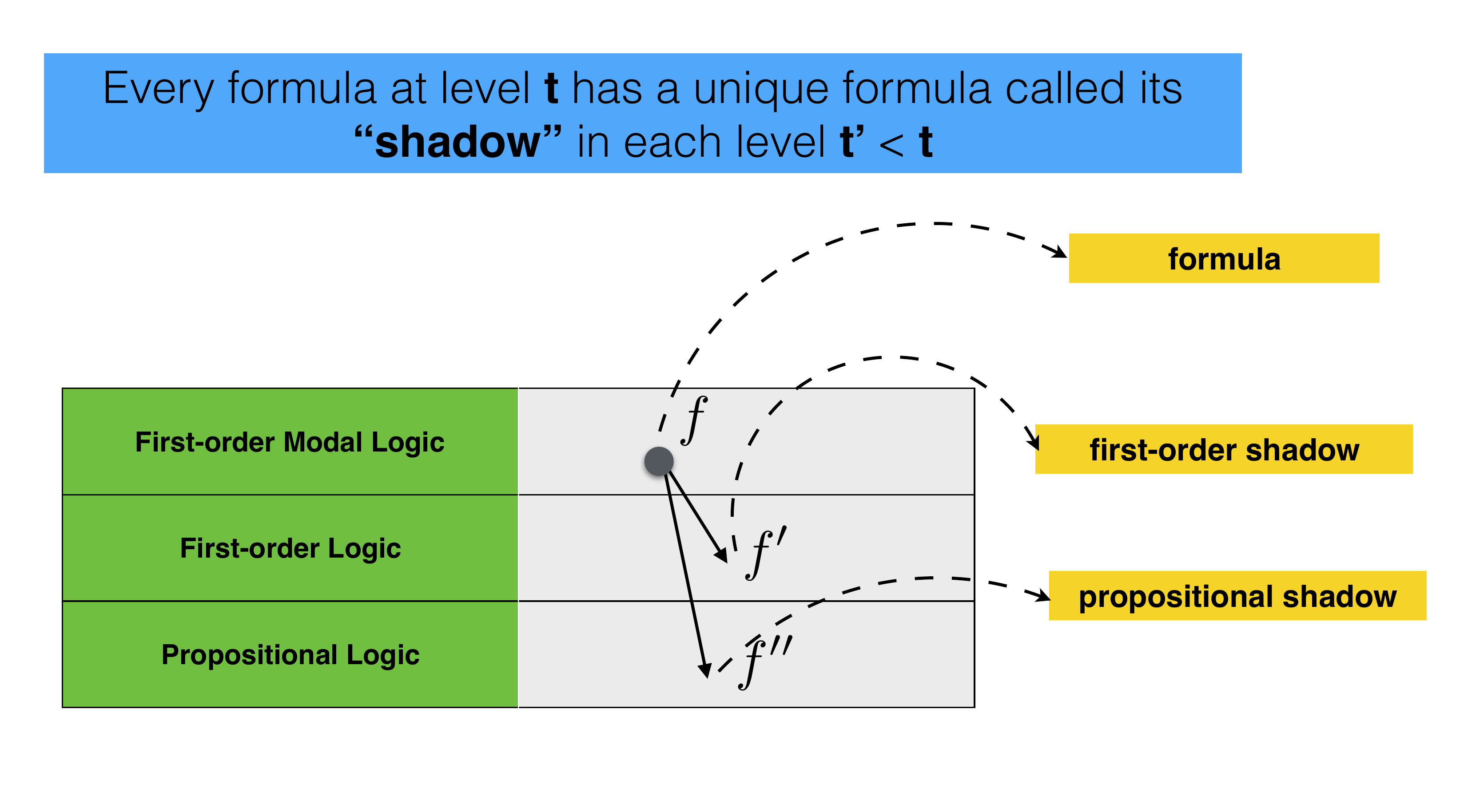}}
      \caption{Shadowing a formulae to different levels (Overview)}
      \label{fig: shadowProver1}
    \end{subfigure}
    \begin{subfigure}[b]{0.45\textwidth}
      \centering
      \shadowbox{
        \includegraphics[width=0.80\linewidth]{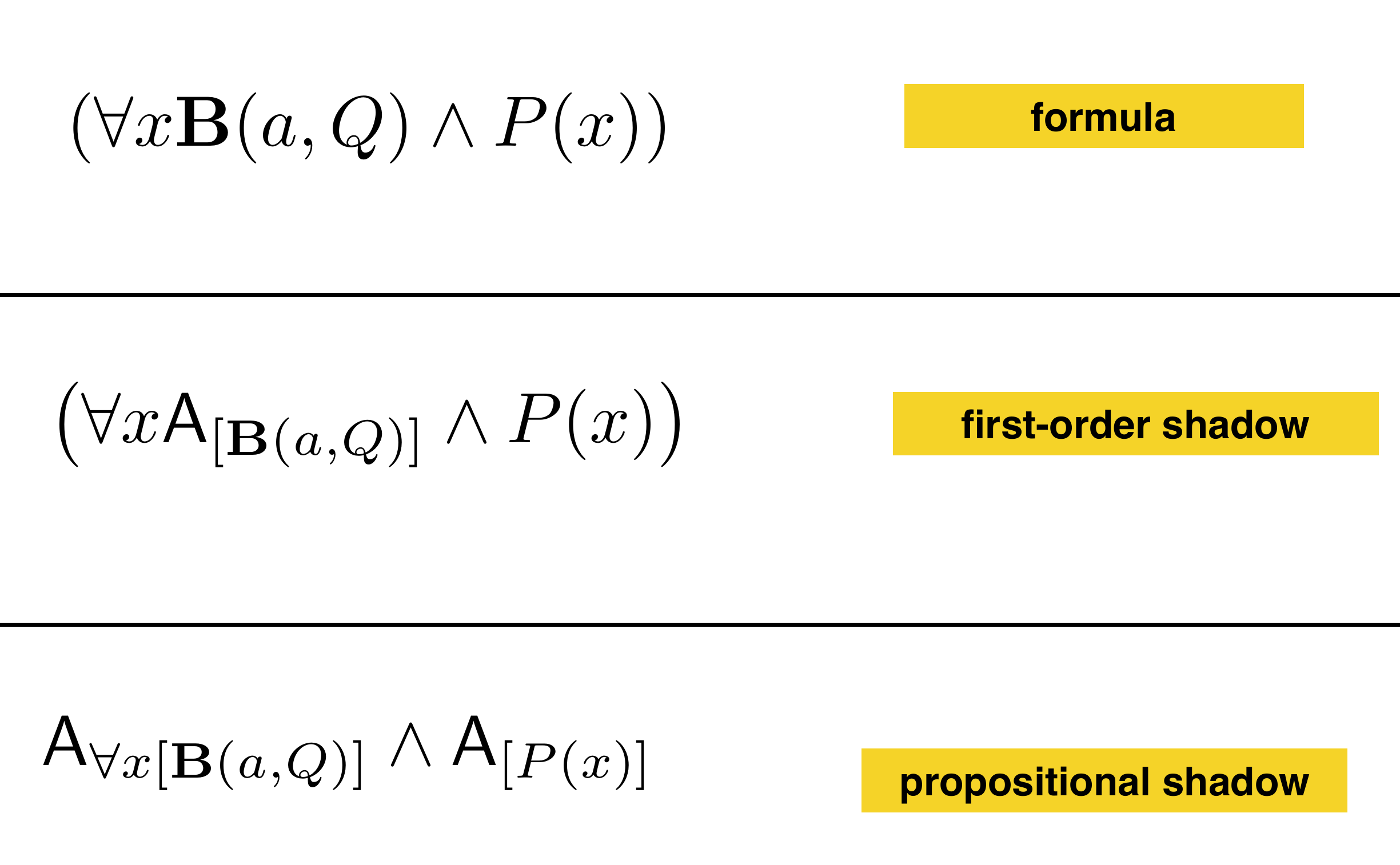}}
      \caption{Shadowing a formulae to different levels (Examples)}
      \label{fig: shadowProver2}
    \end{subfigure}
 \end{figure}

First we define the syntactic operation of \textbf{atomizing} a
formula, denoted by $\mathsf{A}$.  Given any arbitrary formula $\phi$,
$\mathsf{A}_{[\phi]}$ is a unique atomic (propositional) symbol.
Next, we define the \textbf{level} of a formula: $\mathsf{level}:
\Boolean \rightarrow \mathbb{N}$.

 \begin{equation*}
 \begin{aligned}
\mathsf{level}(\phi) = \left\{ 
\begin{aligned}
 &0 ; \phi  \mbox{ is purely propositional
formulae; e.g. } \mathit{Rainy} \\
&1 ; \phi  \mbox{ has first-order
predicates or quantifiers e.g.\ } \mathit{Sleepy}(\mathit{jack}) \\
&2 ; \phi  \mbox{ has modal formulae e.g. } \knows(\mathit{a}, t, \mathit{Sleepy}(\mathit{jack}))
 \end{aligned}
\right.
  \end{aligned}
\end{equation*}
 Given the above definition, we can define the operation of
 \textbf{shadowing} a formula to a level.  See Figures~\ref{fig:
   shadowProver1} and~\ref{fig: shadowProver2}.

  \begin{mdframed}[linecolor=white, frametitle=Shadowing,nobreak,
  frametitlebackgroundcolor=gray!15, backgroundcolor=gray!05,
  roundcorner=8pt]
  To shadow a formula $\chi$ to a level $l$, replace all sub-formulae
  $\chi'$ in $\chi$ such that $\mathsf{level}(\chi')>l$ with
  $\mathsf{A}_{[\chi']}$ simultaneously.  We denote this by
  $\mathsf{S}[\phi,l]$. 

\noindent For a set $\Gamma$, the operation of
  shadowing all members in the set is simply denoted by $\mathsf{S}[\Gamma,l]$.
\end{mdframed}
 
  Assume we have access to a first-order prover $\mathbf{P}$.  For
  a set of pure first-order formulae $\Gamma$ and a first-order
  $\phi$, $\mathbf{P}(\Gamma, \phi)$ gives us a proof of
  $\Gamma\vdash\phi $ if such a first-order proof exists; otherwise
  \color{red}\textbf{fail} \color{black} is returned.  See the
  algorithm sketch given below for a reasoner for $\cc^1$:

  \begin{mdframed}[linecolor=white, backgroundcolor=gray!02,
  roundcorner=8pt]

\begin{footnotesize}

\begin{algorithm}[H]
 \KwIn{Input Formulae $\Gamma$, Goal Formula $\phi$}
 \KwOut{A proof of $\Gamma\vdash\phi$ if such a proof exists,
   otherwise \color{red}\textbf{fail}}
 initialization\;

 \While{goal not reached}{
$ \mathit{answer}= \mathbf{P}\big(\mathsf{S}[\Gamma,1], \mathsf{S}[\phi,1]\big)$\;

\eIf{$\mathit{answer} \not=\color{red}\mathbf{fail}$}{
  \color{blue}{return} \color{black}$\mathit{answer}$ \; }{
  $\Gamma' \longleftarrow$ \textbf{expand} $\Gamma$ by using any
  applicable modal inference schemata\; $\Gamma' \longleftarrow$ \textbf{expand}
  $\Gamma'$ by recursively reasoning forward in all modal contexts\;
  \eIf{$\Gamma' = \Gamma$}{ \tcc{The input cannot be expanded
      further}\ \color{blue}{return} \color{black}
    \color{red}\textbf{fail}} {set $\Gamma \longleftarrow \Gamma'$} }
}
 \end{algorithm}
\end{footnotesize}

\end{mdframed}

%%++++++++++++++++++++++++++++++++++++++++++++++++++++++++++++++++++++

   The algorithm alternates between applying $\{I_{R}\}$ and $\{I_B,
   I_O\}$.  The algorithm is instantiated recursively to handle nested
   first-order reasoning within modal contexts as required in $I_B$.

 \begin{comment}

 \paragraph{{Algorithm II}} The proof calculus for
 $\DCEC$ can be considered to be extension of standard first-order
 proof calculus under different \emph{modal contexts}.  For example, if
 $a$ believes that $b$ believes in a set of propositions $\Gamma$ and
 $\Gamma\vdash_{FOL} Q$, then $a$ believes that $b$ believes $Q$.  We
 convert
 $\believes\left(a, t_a, \believes\left(b, t_b, Q\right)\right)$ into
 the pure first-order formula $Q\left(context(a, t_a, b, t_b)\right)$
 and use a first-order prover. {The conversion process is a bit more
   nuanced as we have to convert compound formulae within iterated
   beliefs into proper conjunctive normal form clauses.

 \end{comment}

 % \subsection{Example Walkthrough}
% \ldots
 
%%% Local Variables:
%%% mode: latex
%%% TeX-master: "main"
%%% End:

\section{Implementation}
\begin{figure}
 \centering
  \includegraphics[width=0.45\linewidth]{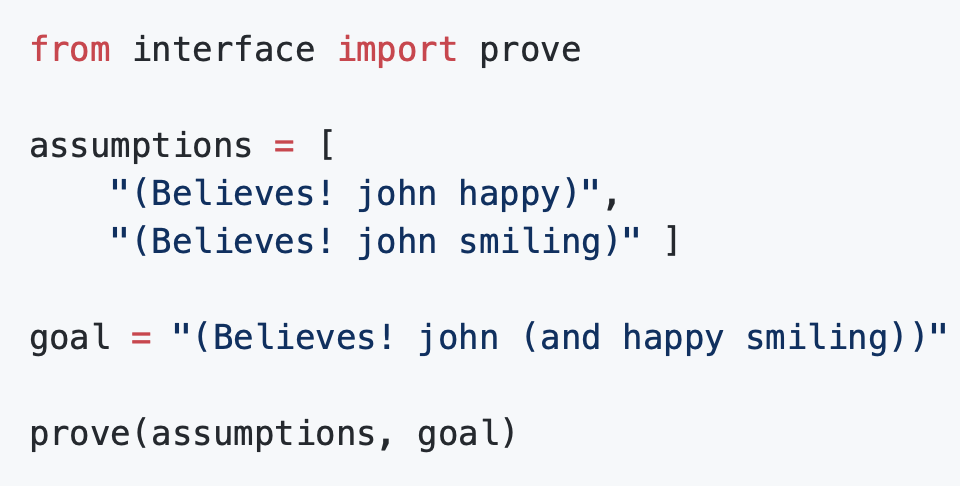}
 \caption{Use of the \DCEC\ reasoner from Python.}
 \label{fig:py}
\end{figure}

\begin{figure}
 \centering
\shadowbox{
  \includegraphics[width=0.6\linewidth]{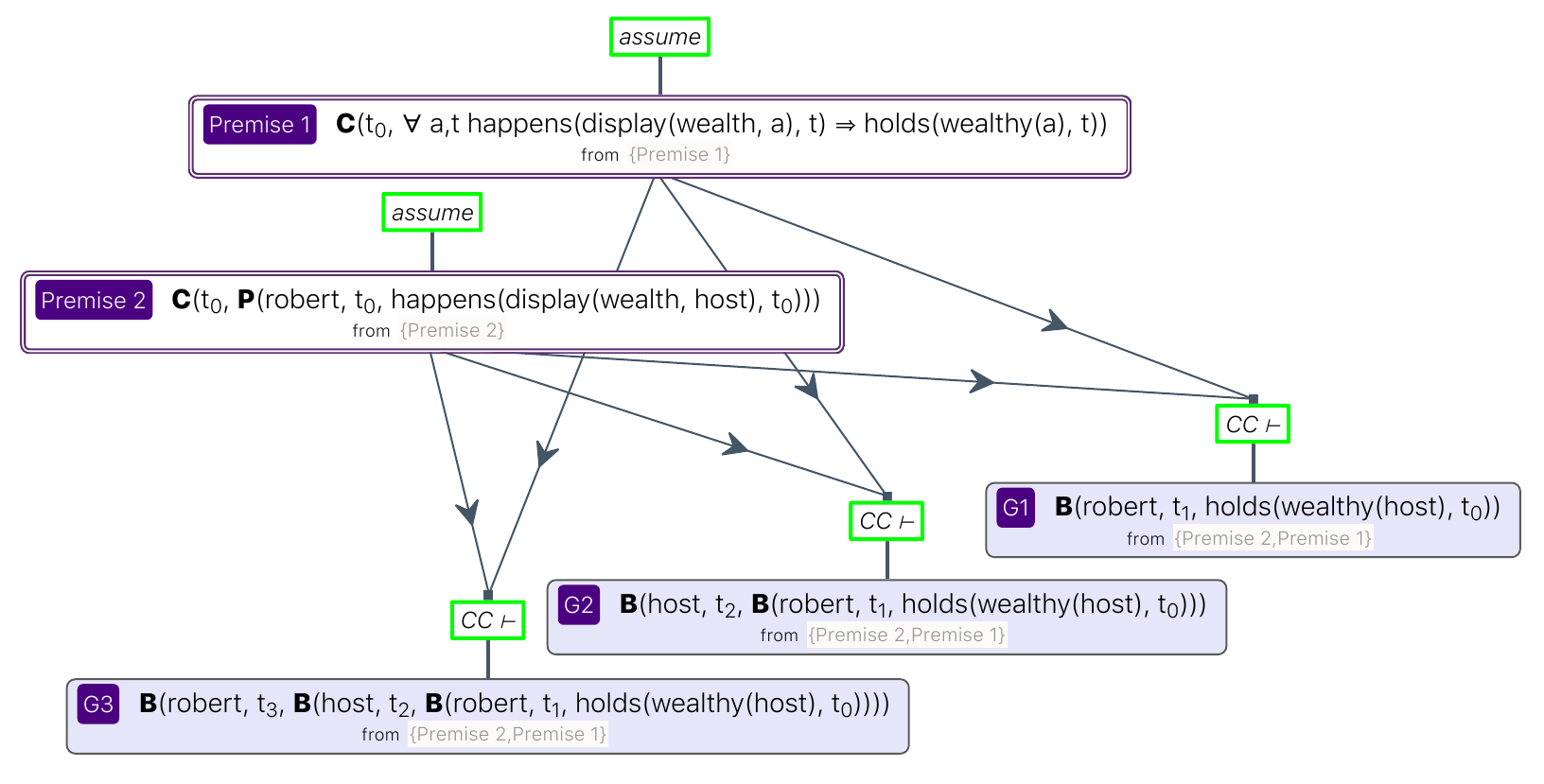}}
 \caption{Use of the \DCEC\ theorem prover within the HyperSlate
   workspace.  Example from \cite{DBLP:journals/corr/abs-1903-03515}.}
 \label{fig:dpex}
\end{figure}

The reasoner is available as an open-source Java library
\cite{naveen_sundar_g_2018_1451808}.  A lightweight Python interface
is available for quick prototyping and experimentation; see
Figure~\ref{fig:py}.  For the first-order prover, we use SNARK, due to
its facilities for extension with procedural attachments and rewrite
systems \cite{snark.94.cade}.  In addition, SNARK comes with theories
for reasoning about simple arithmetic, lists, etc.  For future work,
we shall investigate and pursue integration with other theorem
provers.  Our prover is also integrated within the HyperSlate proof
assistant, a modern extension of the Slate proof assistant
\cite{Slate_at_CMNA08}; see Figure~\ref{fig:dpex} for an example.

%%++++++++++++++++++++++++++++++++++++++++++++++++++++++++++++++++++++
%%++++++++++++++++++++++++++++++++++++++++++++++++++++++++++++++++++++

%%++++++++++++++++++++++++++++++++++++++++++++++++++++++++++++++++++++
%%++++++++++++++++++++++++++++++++++++++++++++++++++++++++++++++++++++

%%% Local Variables:
%%% mode: latex
%%% TeX-master: "main"
%%% End:

\section{Conclusion and Future Work}

We have presented requirements that modal logics for modeling ethical
theories should satisfy. A reasoning algorithm that can satisfy some of
the requirements was presented. Future work involves extending the
reasoner to satisfy the other remaining requirements and proving that
algorithm are sound and complete with respect to a
core inference system. As there are no similar reasoning systems for
\DCEC, direct comparison with other modal logic reasoners is not
possible, but we plan to isolate fragments of \DCEC\ that can enable
benchmarks and comparisons with reasoners for other similar logics..

%%% Local Variables:
%%% mode: latex
%%% TeX-master: "main"
%%% End:

\bibliographystyle{eptcs}
\bibliography{naveen,main72}
\end{document}